\documentclass[11pt,a4paper]{article}
\PassOptionsToPackage{breaklinks}{hyperref}
\usepackage[hyperref]{acl2020}
\usepackage{tipa}

\usepackage{times}
\usepackage{latexsym}
\usepackage{graphicx}
\usepackage{gb4e}
\usepackage{tablefootnote}
\usepackage{enumitem}   
\usepackage[utf8]{inputenc}
\noautomath

\usepackage{todonotes}
\usepackage{microtype}

\aclfinalcopy % Uncomment this line for the final submission
\def\aclpaperid{28} %  Enter the acl Paper ID here

\title{Modelling Verbal Morphology in Nen}

\newcommand{\anu}{\normalfont \Omega}
\newcommand{\unimelb}{\normalfont \mu}
\newcommand{\coedl}{\normalfont \Phi}

\author{Saliha Muradoğlu$^{\anu}$$^{\coedl}$ Nicholas Evans$^{\anu}$$^{\coedl}$ Ekaterina Vylomova$^{\unimelb}$\\
  $^{\anu}$The Australian National University (ANU) ~\;~$^{\unimelb}$The University of Melbourne\\
  $^{\coedl}$ARC Centre of Excellence for the Dynamics of Language (CoEDL) \\
  \texttt{saliha.muradgolu@anu.edu.au},~\;~ \texttt{nicholas.evans@anu.edu.au}, \\
 \texttt{ekaterina.vylomova@unimelb.edu.au}
}
\date{}
\usepackage{gb4e}

\begin{document}
\maketitle
\begin{abstract}

Nen verbal morphology is remarkably complex; a transitive verb can take up to $1,740$ unique forms. The combined effect of having a large combinatoric space and a low-resource setting amplifies the need for NLP tools. Nen morphology utilises distributed exponence -- a non-trivial means of mapping form to meaning. In this paper, we attempt to model Nen verbal morphology using state-of-the-art machine learning models for morphological reinflection. We explore and categorise the types of errors these systems generate. Our results show sensitivity to training data composition; different distributions of verb type yield different accuracies (patterning with E-complexity). We also demonstrate the types of patterns that can be inferred from the training data through the case study of syncretism. 

\end{abstract}

\section{Introduction}
A long-standing research direction in NLP targets the development of robust language technology applicable across the wide variety of the world's languages. Unfortunately, the vast majority of machine learning models are being developed for a small fraction of nearly 7,000 languages in the world, such as English, German, French, or Chinese. 
With introduction of highly multilingual corpora such as UniversalDependencies \cite{nivre2016universal} and UniMorph \cite{sylak-glassman-etal-2015-language,kirov2018unimorph} the situation started to change. For instance, SIGMORPHON organized a number of shared tasks on morphological reinflection starting from 10 languages in 2016 \cite{cotterell-etal-2016-sigmorphon} and up to 90 languages in 2020 \cite{vylomova-etal-2020-sigmorphon}. 
In 2020, languages were sampled from various typologically diverse families:  Indo-European, Oto-Manguean, Tungusic, Turkic, Niger-Congo, Bantu, and others. Still, just one language, namely, Murrinh-patha, an Australian Aboriginal language \cite{mansfield2019murrinhpatha}, represented the whole linguistic variety of the Oceania region.
In this paper, we aim at filling the gap by exploring Nen, a Papuan language spoken by approximately 400 people in Papua New Guinea.  
Nen is known for its rich verbal morphology, with a transitive verb inflecting for up to $1,740$ feature combinations. Distributed exponence, the phenomenon which gives rise to this large paradigm size, provides insight into modelling complex mappings between surface forms and feature bundles. 

We conduct a series of experiments on morphological reinflection task recently introduced under the umbrella of SIGMORPHON \cite{cotterell-etal-2016-sigmorphon,cotterell-etal-2018-conll}. We train several state-of-the-art machine learning models for verbal inflection in Nen and provide an extensive error analysis. We investigate the relationship between the distribution of verb type (inflection classes) in the data and performance. Finally, we show that the system learns properties of the data that are not explicitly given, but may be inferred. 

The rest of the paper is organized as follows: In \hyperref[sec:2]{Section 2}, we give a brief overview of related work. \hyperref[sec:3]{Section 3} provides an overview of Nen verbal morphology, \hyperref[sec:4]{Section 4},
details our methodology, and \hyperref[sec:5]{Section 5} presents our results. Finally, \hyperref[sec:6]{Section 6} concludes the paper.

\section{Related Work}
\label{sec:2}

\citet{muradoglu-etal-2020-compress} is the only reported work on the computational modelling of the Nen language. Similar to this study, the main focus is on modelling Nen verbal morphology, but using finite-state architecture instead. The accuracy achieved by the FST system is 80.3$\%$ obtained across the corpus, with approximately 10$\%$ of the accuracy attributable to the modelling of prefixing verbs (the regularity of copula verbs boosts the accuracy from 70.5$\%$). The accuracies reported are not directly comparable with those presented here due to the different data splits, and increased amount of data.  

In our error analysis, we follow the error taxonomy proposed by \citet{gorman-etal-2019-weird} upon a detailed analysis of typical errors produced by morphologically reinflection systems. A similar study was conducted for Tibetan \cite{di-etal-2019-modelling}.  

\section{The Nen Language}
\label{sec:3}
Nen is a Papuan language of the Morehead-Maro (or Yam) family, located in the southern part of New Guinea \cite{Evans2017}. It is spoken in the village of Bimadbn in the Western Province of Papua New Guinea, by approximately 400 people, for which it is a primary language \cite{Evans2015, Evans2020}. Most inhabitants are multilingual, typically speaking several of the neighbouring languages. 

The subject of this paper -- verbs -- are the most complicated word-class in Nen \cite{Evans2015, Evans2019}. They are demarcated into three separate categories: prefixing, middle, and ambifixing verbs. The latter two are mostly regular in terms of morphophonological rules. In the remainder of this section, we elaborate on these characteristics, to give the reader enough background to follow the discussion in subsequent sections. 

\subsection{Verbal morphology}

We begin our description from the maximal case -- transitive \emph{\textbf{ambifixing}} verbs. Examples of this verb type include \textit{yis} `to plant’ and \textit{waprs} `to do' These verbs allow for full prefixing and suffixing possibilities. \citet{Evans2016} provides the canonical paradigms for the undergoer prefixes, thematics and desinences. Suffix combinations are constructed by concatenating the corresponding thematic and the desinence. Between the undergoer prefix and verb stem is a directional prefix slot, available for all verb types. This slot is occupied by \{\textit{-n-}\}\footnote{We follow linguistic convention with `\{\}' denoting morphemes, and examples are italicised.} to convey a `towards’, \{\textit{-ng-}\} for ‘away’ or left empty to convey a directionally neutral semantic.

\emph{\textbf{Middle}} verbs such as \textit{owabs} `to speak' or \textit{an\={g}s} `to return', are also ambifixing, but the prefixal slot is restricted to \{\textit{n-}\} ($\alpha$--series), \{\textit{k-}\} ($\beta$--series), \{\textit{g-}\} ($\gamma$--series). These prefixes are person and number invariant, and mark the verb as being a dynamic monovalent verb. The prefix set is divided through the use of arbitrarily labels: $\alpha$, $\beta$, and $\gamma$. These dummy indices do not carry specific semantic values until they are unified with other TAM (Tense, Aspect, and Mood) markings on the verb \cite{Evans2015}.

\emph{\textbf{Prefixing}} verbs have separate closed paradigms, tailored to the subtype. Prefixing verbs are mostly distinguished through semantics; positional verbs such as \textit{kmangr} `to be lying down', the verb `to own/have' \textit{awans}, the verb `to walk' \textit{tan} and the copula verb \textit{m} with its directional variants (be hither (i.e. come) or be thither (go)).

Inflectional prefixes for these verbs, mostly resemble the process with ambifixing verbs, yet the suffixes are limited. Of the 50 or so prefixing verbs, the vast majority are positional \cite{Evans2020}. An additional distinguishing feature of prefixing verbs, is the lack of infinitives. Both ambifixing and middle verbs form infinitives through suffixing \textit{-s} to the verb stem. For the purposes of this study, we have listed the prefixing verb lemmas as the verb stem. 

Methodologically, it is more convenient to segment a word as a classical bijective mapping between form to meaning. However, the Nen verbal system distributes information in a more complicated way. The prefixes (undergoer and future imperative) and suffixes (thematic and desinence) are not independent values. Nen verbal morphology is characterised by \emph{distributed exponence (DE)}; ``morphosyntactic feature values can only be determined after unification of multiple structural positions'' \cite{Carroll2016}.

There are two consequences for morphological parsing:

\begin{enumerate}[label=\alph*)]
	\item Provisional unspecified values occur regularly, whether
    \begin{enumerate}[label=(\roman*)]
    	\item These involve partial specification that will be filled in later in the word-parse, such as the left-edge prefix \{\textit{yaw-}\} (2\textbar3 person non-singular undergoer), which will only be made more precise in its number value (dual, or plural) when the thematic is encountered after the verb stem: thus \textit{yaw-aka-ta-n} `I see them\footnote{Can also mean `I see you (more than two)’, resolved by combining with an appropriate free pronoun, \textit{bm} `you (absolutive)’, but for present purposes we ignore this further complication.} (more them two)', where the `non-dual’ marker \{\textit{-ta-}\} eliminates the dual (them two) but \textit{yaw-akae-w-n} `I see them (two)’, where the `dual thematic’\{\textit{-w-}\} eliminates the plural (them more than two) reading.
        \item These involve semantically-unspecified prefix series which only acquire meaning when they are combined with suffixes at the other end of the word: thus \{\textit{yaw-}\}, in the above example, belongs to the $\alpha$-series which, if it combines with the `basic imperfective’, will be given a (broadly) non-past reading, but when it combines with the `past perfective’ it will be given a past reading and when it combines with a ‘projected imperative’ it will be given a future meaning; a $\beta$-series form like \{\textit{taw-}\}, by contrast, will have a `yesterday past’ interpretation when combining with the `basic imperfective’ suffixes but when combining with imperatives it will have a `now/immediate command’ meaning
    \end{enumerate}
    \item More problematically, prefixes that normally have one reading (such as the yaw-example just discussed, which normally marks second/third-person non-singular objects) sometimes have to be given a different meaning (e.g. large plural intransitive subjects) if further parsing to the right encounters a ‘middle’ rather than a ‘transitive dynamic’ stem (Evans 2017, 2019). 
    
    In principle that this means left-to-right morphological parsing is sometimes non-monotonic (particularly in the case of (b)), so that semantic values, as parsing proceeds, need to be sometimes held as provisionally unspecified, sometimes as partially specified, and sometimes as specified but subject to later override. 
\end{enumerate}

\subsection{Distributed Exponence}
\label{sec:Distributed Exponence}

One of the primary motivations for choosing Nen as a case study is the phenomenon that gives rise to this combinatorial power: distributed exponence. Essentially distributed exponence is a morphological phenomenon that gives rise to some types of non-monotonicity.

In linguistics, the notion of extended exponence was first introduced by \citet{mathews1974morphology} and is now commonly referred to as multiple exponence (ME). Matthews defined ME as a category that would have exponents in two or more distinct positions. Distributed exponence is a kind of ME, which involves the use of more then one morphological segment to convey meaning. It requires all relevant morphs to yield a precise interpretation of the feature value in question \cite{Carroll2016, harris2017multiple}.

\begin{exe}
\ex 
\gll n-ng-owan-t-e\\
\small{M:$\alpha$-VEN-set.off-ND:IPF.NP-IPF.NP.2\textbar3SGA}\\
\trans `You/(s)he are/is setting off.'\footnote{Example adapted from \cite{Evans2020}}
\end{exe}

In the example above, no one marker marks the singular person. The information of the agent being singular is distributed across the thematic (dual/non-dual) and the desinence (single/dual/plural). If a non-dual thematic is present than the desinence cannot have dual features; the only options are singular or plural. Another morpheme present in this example is the prefix {-ng-} which marks the verb with the directional \textit{thither}. The prefix {n-} marks this verb as a middle verb; it reduces the valency of the verb and yields information about the membership of the class $\alpha$. Together with the prefix, thematic and desinence, the TAM feature can be obtained. 

\section{Methodology}
\label{sec:4}

\subsection{Morphological reinflection task}
Morphological inflection is a task of predicting a target word form from a corresponding word lemma and a set of morphosyntactic features (specifying the target slot, e.g. its part of speech (POS), tense, number, gender). For instance, a system is provided with a lemma ``to sing'' and a set of tags ``Verb; Past'' and needs to generate ``sang''.
Morphological \emph{reinflection} is a variation of the task when a lemma form is replaced with some other form and (optionally) its tags. The task has been traditionally solved with finite-state transducers, either hand-engineered \cite{koskenniemi1983two,kaplan1994regular} or trainable models that rely on both expert knowledge and data \cite{mohri1997finite,eisner2002parameter}. In 2016 SIGMORPHON started a series of shared tasks on morphological reinflection, and neural models demonstrated superior performance when compared to finite-state or rule-based approaches, especially in high-resource languages
 \cite{cotterell-etal-2016-sigmorphon,vylomova-etal-2020-sigmorphon}.

\subsection{Data}
The data used in this study comes from a Nen verb corpus (approximately $6,000$ verb samples representing $2,231$ unique inflected forms) created by \citet{Muradoglu2017}. This dataset is a distilled subset from the approximately 8-hour natural speech corpus for the Nen language. As such it entails a frequency sorted list of all the verb forms occurring.

The training data is a set of triples comprising a lemma, morphosyntactic features, and an inflected form (i.e. we will only focus on morphological \emph{inflection}). 

\paragraph{Sampling}

Following the methodology in \citet{cotterell-etal-2018-conll} we split the data into training, development, and test sets. Training splits were created by sampling without replacement for three set sizes: all (\textit{ALL}), medium (\textit{MR}), and low (\textit{LR}). 
 
In virtue of coming from a natural corpus, the list of verb forms we use is Zipfian. This study does not distinguish between the feature bundles and only considered surface (inflected) forms. To facilitate the nature of our study, we uniformly distribute frequency across each syncretic cell. 

For the \textit{ALL} training set we start by sampling the first $1,931$ forms, in accordance with the Zipfian ranking across the corpus. In other words, we sample the $1,931$ most frequent verb forms. We randomly shuffle the remaining $300$ forms into a $200$ form test, and $100$ form development (dev) sets. The test and dev sets remain the same through this experiment. Zipfian sampling is considered more realistic in this case, as it mimics the stimulus a language learner encounters. The dev and test set are randomly shuffled since supervised methods usually generalise from frequently encountered words.  

For the \textit{LR} and \textit{MR} settings we take the first $100$ and $1,000$ forms from the \textit{ALL} training set, respectively.
In addition, we create a high-resource (\textit{HR}) set by supplementing the \textit{ALL} set with synthetic forms, the final set contains $10,000$ forms. In order to generate synthetic samples, we use data hallucination technique proposed in  \citet{anastasopoulos19emnlp}. Note that the  low-resource (\textit{LR}) training set is a subset of the medium-resource (\textit{MR}), which is supersetted by the \textit{ALL} (and by extension the high-resource (\textit{HR}) data set). 

Finally, we contrast Zipfian sampling, when forms are sampled based on their frequency, to random sampling. Both sets (\textit{LR} and \textit{MR}) for the random sampling are created in a similar manner to Zipfian sampling, except frequency is not considered.  Note that due to initial data size constraints, the \textit{ALL} (and, therefore, \textit{HR}) data sets for \emph{both} the Zipfian and random sampling are the same. \footnote{Since the test and dev set are the same for both sampling methods, and are generated from the \textbf{remaining} $300$ tokens (i.e. the least frequent items), it renders the random sampling of the \textit{ALL} (and thus \textit{HR}) the same. } 

\subsection{Experiments}
In the current study we conducted three experiments to address our research questions. 

\subsubsection{Experiment 1: Testing across various data sizes and sampling methods}
\label{exp1}
\textit{Research Question: How does training size and sampling method affect the models' performance, and what kind of errors are likely across these conditions?}

We evaluate modelling accuracies across four different training sizes, which is further contrasted across sampling type. Our experimental setup mirrors those of the SIGMORPHON reinflection tasks \cite{cotterell-etal-2016-sigmorphon,cotterell-etal-2017-conll, cotterell-etal-2018-conll,vylomova-etal-2020-sigmorphon}: given an input lemma and a set of feature tags, models generate inflected forms. The final accuracy is computed as the percentage of matches between the gold and predicted forms.

\subsubsection{Experiment 2: Testing compositionality of training data} 
\label{exp2}
\textit{Research Question: Does the composition of the training data affect the resultant accuracies, and, if so, how?}

We test the effects of the verb type composition (i.e. how much of each verb type there is) in the training set.  This study consists of seven (arising from all combinations of the three
verb types) training data sets obtained through the sampling methods outlined above. We compare training sets of ambifixing verbs only, prefixing verbs only, middle verbs only, a two-way combination of each verb class: ambifixing and prefixing verbs, ambifixing and middle verbs, and prefixing and middle verbs and, finally an equal distribution of all three verb types, as listed in Table~\ref{comp-table}. Each set contains 386 forms (instances), stipulated by the amount of prefixing verbs available.  The test and development set are 100 forms each, and is made up of 34 ambifixing, 33 middle and 33 prefixing verbs \footnote{Uniform distribution is unlikely in natural language, in fact, \citet{Muradoglu2017} shows that the distribution is skewed to favour a higher number of ambifixing verbs in terms of the number of inflected forms.}

\subsubsection{Experiment 3: Testing syncretism}
\label{exp3}
\textit{Research Question: Do the models infer properties of the language which are not annotated in the data?} 

In Nen, the second and third-person feature bundles often correspond to the same surface form across the available TAM categories (i.e. are syncretic). We test the likelihood of both models predicting the \textit{unseen} second-person singular for the past perfective TAM category as syncretic with the \textit{seen} third-person singular variant. This is the one instance across the Nen verbal paradigm where this syncretism does not hold. In essence, we examine linguistic patterns that may be inferred from an annotated dataset. 

The main focus here, is to categorise the type of prediction rather than the overall accuracy, as such training and development sets are identical to those generated for the \textit{ALL} setting in the first experiment. The test set is comprised of 100 inflections of the past perfective second singular tags, most of these have been gathered from the Nen dictionary \cite{dictionaria-nen}.

\subsection{Models}
For our experiments, we will utilise two models that have shown superior performance in SIGMORPHON--CoNLL 2017 Shared Task on morphological reinflection in low- and medium-resource settings \cite{cotterell-etal-2017-conll}. Both of them are essentially neural sequence-to-sequence models implemented in Dynet \cite{neubig2017dynet}. 
In addition, we also compare the results with a simple non-neural baseline used in 2017--2018 tasks on morphological reinflection \cite{cotterell-etal-2017-conll,cotterell-etal-2018-conll}.

\paragraph{Hard Monotonic Attention  \cite{aharoni-goldberg-2017-morphological}}
An external aligner \cite{sudoh2013noise} first produces transformation operations between an input (lemma) and a target (inflected form) character sequences. The alignment operations (steps) are then fed into a neural encoder--decoder model. The network, therefore, is trained to mimic the transformation steps, and at inference time it predicts the actions based on the input (lemma) sequence.   
Unlike soft attention models, this model attends to a single input state at each step and either writes a symbol to the output sequence or advances its pointer to the next state. Hard attention models demonstrate superior performance in languages that employ suffixing morphology with stem changes.   

\paragraph{Neural Transition-based \cite{makarov-clematide-2018-neural}}
The model is essentially derived from \citet{aharoni-goldberg-2017-morphological} by enriching it with explicit insertion, deletion or, alternatively, copy mechanisms. The copy mechanism led to significant accuracy gains in low-resource settings. Following \citet{rastogi-etal-2016}, the model can be seen as a neural parameterization of a weighted finite-state machine.

\paragraph{Non-neural Baseline \cite{cotterell-etal-2017-conll,cotterell-etal-2018-conll}}
The non-neural system first aligns lemma and inflected form strings using Levenstein distance \cite{levenstein1966binary} and then extracts prefix- and suffix-based transformation rules. 

\subsection{Settings}
The hyperparameters of the models are set to the values reported in the corresponding papers as per Table~\ref{hype-table}. 

\begin{table}[h]
\centering
\begin{tabular}{l|r|r}
\hline \textbf{Hyperparameters} & \textbf{A\&G}& \textbf{M\&C}\\ \hline
Input dim & $100$& $100$\\
Hidden dim& $100$& $100$\\
Epochs & $100$& $50$\\
Layer & $2$ & $1$ \\
\hline
\end{tabular}
\caption{\label{hype-table} Hyperparameters for both A\&G (2017) and M\&C (2018) models.}
\label{table}
\end{table}

\section{Results}
\label{sec:5}
Table~\ref{acc-table} shows the accuracies achieved for each system for each training set size and sampling type from \hyperref[exp1]{Experiment 1}. For all setups the M\&C model performed best with random sampling (where applicable). As expected the high-resource setting performs best overall. The random sampling yields slightly higher accuracies than the Zipfian counterpart, this is likely due to the fact that prefixing verbs, particularly the copula and its 40 distinct forms occupy a majority of the top 100 positions in the Zipfian distribution. Thus when random sampling is utilized the training set includes more examples of ambifixing verbs. 

\begin{table*}[!h]
\centering
\begin{tabular}{r|r|r|r|r|r|l}

                   & \multicolumn{2}{c|}{\textbf{A\&G 2017}} & \multicolumn{2}{c|}{\textbf{M\&C 2018}} & \multicolumn{2}{c}{\textbf{Non-Neural baseline (NNB)}} \\ \cline{2-7} 
                   & Random        & Zipf         & Random        & Zipf         & Random               & Zipf              \\ \hline

HR                 & \multicolumn{2}{c|}{0.610}       & \multicolumn{2}{c|}{\textbf{0.650}}           & \multicolumn{2}{c}{0.015}             \\ \hline
ALL                & \multicolumn{2}{c|}{0.390}      & \multicolumn{2}{c|}{\textbf{0.510}}          & \multicolumn{2}{c}{0.010}         \\ \hline
MR                 & 0.295        & 0.285        &  \textbf{ 0.445 }        &  0.420           &     0.000            & 0.000                  \\ \hline
LR                 & 0.020             & 0.005            &    \textbf{0.080}         &    0.030        &  0.010         &     0.010            \\ \hline

\end{tabular}
\caption{\label{acc-table} Data set, model and sampling accuracies. ALL is a total of 1,931 verbs, HR is 10,000, MR is 1,000 and LR is 100 samples for the training set.}
\label{table2}
\end{table*}

\begin{table*}[!h]
\centering
\begin{tabular}{r|r|r|r|r|r|r|r|r|r|r|r|r}
               & \multicolumn{3}{c|}{\textbf{ALL}} & \multicolumn{3}{c|}{\textbf{HR}} & \multicolumn{3}{c|}{\textbf{MR}}                           & \multicolumn{3}{c}{\textbf{LR}}                               \\ 
\cline{2-13}
               & \rotatebox{90}{A\&G } & \rotatebox{90}{M\&C  } & \rotatebox{90}{NNB}  & \rotatebox{90}{A\&G }  &\rotatebox{90}{M\&C  } & \rotatebox{90}{NNB}  &  \rotatebox{90}{A\&G } &\rotatebox{90}{M\&C  } & \rotatebox{90}{NNB}   &  \rotatebox{90}{A\&G }  & \rotatebox{90}{M\&C  }& \rotatebox{90}{NNB}    \\ 
\hline
Allomorphy     & 56      & 55      & 190  & 54      & 46      & 144 & 61      & 77                                & 188 & 17                                   & 162     & 190  \\ 
\hline
Free Variation & 30      & 24      & 0    & 14      & 15      & 11  & 13      & 24                                & 0   & 0                                    & 2       & 0    \\ 
\hline
Target         & 8       & 8       & 8    & 8       & 8       & 8   & 8       & 8                                 & 8   & 8                                    & 8       & 8    \\ 
\hline
Stem           & 28      & 11      & 0    & 2       & 1       & 5   & 61      & 7* & 2   & 174† & 22      & 0    \\ 
\hline
Total          & 122     & 98      & 198  & 78      & 70      & 168 & 143     & 116                               & 198 & 199                                  & 194     & 198  \\
\hline
\end{tabular}
\caption{\label{error-table} Absolute number of errors on the test set (200 instances) made by each system trained in ALL, HR, MR and LR setting.*contains 5 looping errors,† 17 looping errors.}
\label{terror}
\end{table*}

\subsection{Error Analysis}

We analysed the errors produced in prediction following the taxonomy laid out by \citet{gorman-etal-2019-weird,di-etal-2019-modelling}. 

We have taken a hierarchical approach to our error classification; whereby if more than one error is present, the category higher up is reported. For example, if a predicted form exhibits both target and allomorphy errors (error types are described in the following subsections), then only the target error is reported. The motivation for this lies in the nature of the error; free variation is technically not even an error. By contrast, misapplication of a morphophonological rule does indeed yield an incorrect form. Additionally, we have marked Target errors higher up as the system cannot be expected to correctly predict a form if the gold standard is incorrect. The hierarchy is as follows: Target$>$Stem$>$Allomorphy$>$Free Variation.
3
Table~\ref{terror} summarises the types of errors across the different training sizes for each model. Overall, for both systems allomorphy errors remain relatively unimproved between the \textit{ALL} and \textit{HR} setting, but show a leap of reduction from the LR to MR conditions. Free variation errors are more prevalent in the \textit{ALL} setting. This is probably a consequence of seeing more of the golden data and thus observing more of the systematic variations. This also explains why these errors reduce in number for the \textit{HR} setting. The target errors are consistent across each experiment, as these are systematic issues with the gold data. Interestingly, stem errors reduce in the HR setting. This is despite the use of hallucinated data. 

\subsubsection{Allomorphy}

This category consists of errors which are characterised by a misapplication of morphophonological rules, or feature category mappings. Frequent errors include the absence of vowel harmony or place assimilation rules, and incorrect mapping of feature bundles to surface forms. Most errors are of this category.

\textbf{Vowel harmony.} The Nen language exhibits vowel harmony. Consider the form \textit{yn$\bar{g}$ite} generated by one of the models, in a canonical sense the inflection is correct, but the presence of the high front vowel \textit{i} requires the general \textit{e} to harmonize to become \textit{yn$\bar{g}$iti}.

\textbf{Morphophonological Rules.} When combining \textit{r} final stems with \textit{t} phonemes (which occurs in inflections via the non-dual thematics or certain desinences with $\emptyset$ thematics), the resultant sound is \textit{n} \cite{Evans2016}. The M\&C systems predicts that the stem \textit{tar} inflected for the non-prehodiernal, first person actor and third-person undergoer as \textit{ytaretan}. Presumably, the break down is \textit{y-tar-e-ta-n}. Interestingly, it inserts an \textit{e} between the \textit{r} and \textit{t}, rather than concatenates the stem with the \{-ta-n\} suffix. The correct form is \textit{ytanan}.

\textbf{Misapplication of category.} These errors are rather straightforward: they are a misapplication of inflection rule and result in an incorrect cell of a paradigm. For example, \textit{ynrenzan} is generated instead of \textit{ynrenzng}. Technically, the form generated is correct, but it should correspond to the past perfective, first-person singular acting on dual actor suffix. Instead, it is mapped to the imperfective non-prehodiernal, third singular acting on dual actor suffix. 

\textbf{Future Imperatives.} In all settings, across all systems tested, the future imperative was incorrectly predicted. Much like the $\beta$ and $\gamma$ counterparts, the system generated this TAM category by simply choosing an $\alpha$ prefix and suffixing \{-ta\}. Both A\&G and M\&C systems produce \textit{yngita} instead of \textit{yngangwita}. This formulation is correct for the  $\beta$ and $\gamma$ series producing the imperfective imperative and mediated imperative, respectively. However, the future imperative has a special prefix which prefixes after the undergoer and directional prefix. It signifies the future imperative TAM category and marks the agent as either singular or non-singular. 

\textbf{Prefixing Verbs.} Given the sparsity of examples for prefixing verbs and in particular their subtypes, a common occurrence across the data sizes is for the prefixing verb predictions to be inflected with the wrong features. For example, when the verb \textit{m} `to be' is inflected for the andative, 3PL+ undergoer and imperfective non-prehodiernal TAM the correct inflected form would be \textit{yenewelmän}, instead the system gives \textit{ynm} which it correctly identifies as a related form, but it does not have the correct inflectional features.

\subsubsection{Free variation}

Free variation errors occur when more than one acceptable inflected form exists; this is particularly true of the data set used in this study. The corpus used here has been distilled from a natural speech corpus, that has been transcribed. In addition to spelling variation - that arose as the orthographic decisions changed with ongoing documentation, the corpus also exhibits inter-speaker variation. An example includes: \textit{yérniwi} as the predicted form and \textit{yrniwi} as the gold standard. In Nen orthography, epenthetic vowels are not written in as their locations can be predicted \cite{evans2016nen}. Older transcriptions wrote these vowels in with the \textit{é}. 

\subsubsection{Target}

These errors are characterised by incorrect feature tags in the gold standard data. One such example is as follows: the model predicts the form to be \textit{nnganztat} and the gold standard is given as \textit{ynganztat}, the feature tag, however, includes [M]\footnote{[M] marks the verb as middle and is present when one of the three middle prefixes is present.} and not [3SGU]. In such cases, based on the feature bundle, the system generated form is correct. This particular mismatch of middle and transitive verbs is the main source of this kind of error; it arises from the fact that a single verb may have a middle and transitive verb variant. This distinction can be difficult to decipher, and on some occasions, it can even be a result of speaker error.

\subsubsection{Stem}

This category denotes either a generated stem or a re-mapping of a seen but irrelevant stem, to the inflected form. These errors have linguistically viable morphemes attached, but we have not evaluated the accuracies of the mapping between feature and form for the morphemes.

\textbf{Remapping}

One such example is A\&G model generating \textit{ygmtandn} for the stem \textit{sns}. It appears that the \textit{gms} stem has been (incorrectly) inflected and mapped to the feature bundle of the \textit{sns} stem. The correct inflection is \textit{ysnendn}. 

\textbf{Generated Stem}

The less frequent of the two are stems that have been randomly generated. For example with the stem given as \textit{renzas}, the system generates: \textit{ymryawem} in place of \textit{yrenzawem}.

We have also encountered several looping errors such as: 
\textit{ynawemaylmyylmyylmyylmy-
ylmyylmyymayamawemyymamyamawemyymamya-
mawemyymamyamawemyymamyamawemyymamyam-
awemyylmyamyamawemyymamyamawemyymamya-
mawemyylmyylmyy} where the correct form is \textit{ysnewem}.

\subsection{Composition study}

\begin{table}[!h]
\centering
\begin{tabular}{r|p{2.4em}|p{2.4em}|p{1cm}}

  & \textbf{A\&G} & \textbf{M\&C} & \textbf{NNB} \\ \hline
Ambifixing only   & 0.111     & 0.170 & 0.010           \\ \hline
Middle only       & 0.121     & 0.210 & 0.111           \\ \hline
Prefixing only    & \textbf{0.212}    & 0.250 & 0.010           \\ \hline
Ambi + Pre        & 0.111     & 0.190 & 0.010           \\ \hline
Ambi + Mid        & 0.071     & 0.130 & 0.040           \\ \hline
Mid + Pre         & 0.141     & \textbf{0.290} & 0.040           \\ \hline
Ambi + Mid + Pre  & 0.061     & 0.200 & 0.040        \\ \hline

\end{tabular}
\caption{\label{comp-table} Data sets for each composition type, model and sampling accuracies. The training size for each is $386$ forms (defined by the available prefixing verbs).}
\label{table3}
\end{table}

\begin{table*}[!h]
\centering
\begin{tabular}{r|r|r|r|r|r|r}
                 & \multicolumn{2}{c|}{Ambifixing} & \multicolumn{2}{c|}{Middle} & \multicolumn{2}{c}{Prefixing}  \\ 
\cline{2-7}
                 & AG    & MC   & AG    & MC   & AG    & MC   \\ 
\hline
Ambifixing only  & 11 & 15  & 2 & 0  & 0 & 2                  \\ 
\hline
Middle only      & 2 & 1  & 12 & 19  & 0 & 1                  \\ 
\hline
Prefixing only   & 0 & 0  & 0 & 0  & 21 & 24                  \\ 
\hline
Ambi + Pre       & 1& 1  &1 & 0  & 10 & 18                 \\ 
\hline
Ambi + Mid       & 1 & 4  & 6 & 8  & 0 & 1                 \\ 
\hline
Mid + Pre   & 0 &  3  & 3 &  10 & 11 &  16                      \\ 
\hline
Ambi + Mid + Pre & 0 & 6  & 4 &  8  & 3 & 6                     \\
\hline
\end{tabular}
\caption{\label{correct-table} Absolute number of correct predictions for each setup.}
\label{tablec}
\end{table*}

In \hyperref[exp2]{Experiment 2}, we tested the effects of training set composition; in other words, the informative nature of each verb type. 

As mentioned above, the ambifixing verb class has the largest combinatorial space, reducing in size as we consider middle and prefixing verbs, respectively. Another way to consider this would be by providing comprehensive lists of the morphemes in a given language (such as \citet{bickel2005inflectional,shosted2006correlating}). Thus, the complexity of an inflectional system is measured by enumerating the number of inflectional categories and the range of available markers for their realisation (i.e. E-complexity). The bigger the number, the more complex the resulting system is.\footnote{Although more recent works have explored the issues with E-complexity \cite{ackerman2013morphological}, we use it here as a guiding principle and acknowledge that further work is required to make a more nuanced statement. } With this in mind, we would expect that, given the same training size for each verb type, the ambifixing would perform the worst,\footnote{The combinatorial space for a transitive verb is $1,740$ cells \cite{muradoglu-etal-2020-compress}}  then the middle followed by the prefixing verbs. Our results, shown in Table~\ref{table3}, confirm this hypothesis. 

More revealing than the overall accuracy for each set and model combination, is a decomposition of accuracy according to the verb class. Table~\ref{tablec} summarises the performance for each category according to verb class. Unsurprisingly, when the training set contains only one type of verb, it performs best for the type of verb seen in the training data.

From a linguist perspective, with principle parts from the middle verbs (mainly the suffixal system, recall that the middle verb takes a dummy prefix to reduce valency) and prefixing verbs (prefixal paradigm) we can construct the full paradigm available to ambifixing verbs. The results presented here show no such compositionality; instead, we see a simple correspondence to verb type observed. 

As expected, we see the weak \textit{leaking} or overlap between ambifixing and middle verbs, with very little transferability from prefixing to other verb types. It highlights the importance of tag choice; middle verbs have a [M] tag for the undergoer prefix, to mark the dummy prefix. If this tag were absent, would we see more transferability between ambifixing and middle verbs? Linguistically, no information would be lost as the absence of this tag still allows for the middle verbs to be clustered together. 

\subsection{Syncretism test}

\hyperref[exp3]{Experiment 3}, entailed testing the systems with an unseen feature bundle and analysing the predicted forms, to gauge whether the models learnt syncretic behaviour. 

As can be seen by the suffixal paradigm found in \citet{Evans2016},\footnote{Table 23.14 (pg 563) and Table 23.16 (pg 565)} where both numbers are available, almost all the TAM categories exhibit syncretism across the second and third-person singular actor. The past perfective slot is the only case with distinct forms for the second and third singular person numbers. We are testing the prediction of an exception. The second singular is formed with \{-nd-$\emptyset$-\} and the third-person singular with the \{-nd-a\} suffix. We note the similarity between the second singular and dual forms, where the second dual is \{-a-nd\}. This becomes particularly pertinent when a vowel is inserted between consonants for ease of articulation but must also adhere to vowel harmony. In such cases, the second dual and second singular may appear the same. 

Using the \citet{aharoni-goldberg-2017-morphological} architecture, the model incorrectly predicts 81 out of the 100 test forms as the third singular perfective category with the suffix \{-nd-a\} instead of \{-nd--$\emptyset$-\}. Four forms predicted correctly (likely due to the similarity between the surface forms of the second-person dual and singular tags) and the remaining fifteen distributed across second-person dual and plural actor of the same TAM category, second/third singular for the imperfective non-prehodiernal TAM category, and several instances of nonce inflections such as \{-ngt\} or \{-ngw\}. 

Similarly, the \citet{makarov-clematide-2018-neural} system overwhelmingly predicts the unseen second singular form to be syncretic with the third singular (90 out of the 100 forms are predicted as such). Of the remaining ten instances three are correct, four are incorrectly modelled as the imperfective imperative (yet given the prefixing series is $\alpha$, the future imperative prefix is absent) and one of each: second/third imperfective non-prehodiernal, second/third neutral preterite or second dual past perfective. 

From these results, it is clear that such systems not only observe patterns that are directly stipulated through annotation but also others that may be inferred from the data. It is important to note this behaviour, particularly in cases such as the one presented here as the verb corpus only entails two instances of the second singular past perfective. 

\section{Conclusion}
\label{sec:6}

Diversity representation of languages in NLP is vital to test the generalisations of models. We present the first-ever neural network-based analysis of Nen, the first representation of the Yam language family and to the best of our knowledge, of a Papuan language. Nen provides an interesting case study as it exhibits non-monotonic morphological mapping: distributed exponence.

We compare state-of-the-art models for morphological reinflection across various training sizes and two sampling methods: random and Zipfian. The results show no significant difference between sampling methods, and minor differences may be attributed to training set composition differences. In the Zipfian case, the prefixing verb types are over-represented as they are more frequent in natural speech.  
We provide extensive analysis of types of errors generated by each system and show that the most common error type is allomorphy errors; a misapplication of morphophonological rules, or feature category mappings. We introduce a new subcategory of error type: free variation, which is a consequence of the natural speech origins of the corpus. 

 We further explore composition effects by generating training sets with incremental distributions for the three verb classes noted. As expected, we found that the models trained with one class had higher prediction accuracy for that class. Across homogeneous compositions, the prefixing verb class performed the best. This is likely due to a smaller E-complexity -- or more simply -- a smaller combination of feature tags for which the system must learn mappings. Finally, we explore the likelihood of learning syncretic behaviour and using this as a predictor for an unseen feature bundle -- the second singular past perfective. Overwhelmingly, the system incorrectly predicts syncretism with over 80$\%$ for the A\&G system and 90$\%$ for the M\&C system. These results highlight that these systems can infer patterns from the data sets provided. Although in our case the prediction of syncretism mirrors that of a human learner, there may be underlying, unwanted properties learnt from the data given, which calls for careful preparation of data and observation of output. 

\bibliography{acl2020}
\bibliographystyle{acl_natbib}

\end{document}